\documentclass{bmvc2k}

\pdfoutput=1	

\usepackage{amsmath}
\usepackage{amssymb}

\usepackage{booktabs}	

\usepackage[]{units}

\DeclareMathOperator*{\argmax}{argmax}

\title{Hybrid One-Shot 3D Hand Pose Estimation by Exploiting Uncertainties}

\addauthor{Georg Poier}{poier@icg.tugraz.at}{1}
\addauthor{Konstantinos Roditakis\textsuperscript{2,}}{croditak@ics.forth.gr}{3}
\addauthor{Samuel Schulter}{schulter@icg.tugraz.at}{1}
\addauthor{Damien Michel}{michel@ics.forth.gr}{2}
\addauthor{Horst Bischof}{bischof@icg.tugraz.at}{1}
\addauthor{Antonis A. Argyros\textsuperscript{2,}}{argyros@ics.forth.gr}{3}

\addinstitution{
 Institute for Computer Graphics and Vision\\
 Graz University of Technology\\
 Graz, Austria
}
\addinstitution{
 Institute of Computer Science\\
 FORTH\\
 Heraklion, Greece\\
}
\addinstitution{
 Computer Science Department\\
  University of Crete\\
  Heraklion, Greece
}

\runninghead{Poier \etal}{Hand Pose Estimation by Exploiting Uncertainties}

\def\eg{\emph{e.g}\bmvaOneDot}

\def\etal{\emph{et al}\bmvaOneDot}
\def\ie{\emph{i.e}\bmvaOneDot}

\begin{document}

\maketitle

\begin{abstract}
   Model-based approaches to 3D hand tracking have been shown to perform well in a 
wide range of scenarios. However, they require initialisation and cannot recover 
easily from tracking failures that occur due to fast hand motions. Data-driven 
approaches, on the other hand, can quickly deliver a solution, but the results often suffer from lower 
accuracy or missing anatomical validity compared to those obtained from model-based approaches. In this work we propose a hybrid approach for hand pose estimation from a single depth image. First, a learned regressor is employed to deliver multiple initial hypotheses  for the 3D position of each hand joint. Subsequently, the kinematic parameters of a 3D hand model are found by deliberately exploiting the inherent uncertainty of the inferred joint proposals. This way, the method provides anatomically valid and accurate solutions without requiring manual initialisation or suffering from track losses.
Quantitative results on several standard datasets demonstrate that the proposed method outperforms state-of-the-art representatives of the model-based, data-driven and hybrid paradigms. 
\end{abstract}

\section{Introduction}
\begin{figure}[t]
  \centering
  \subfigure[]{\includegraphics[width=0.15\columnwidth]{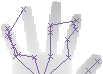}\label{fig:tochange03}}
  \quad
  \subfigure[]{\includegraphics[width=0.15\columnwidth]{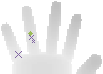}\label{fig:tochange04}}
  \quad
  \subfigure[]{\includegraphics[width=0.15\columnwidth]{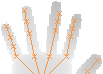}\label{fig:tochange05}}
  \quad
  \subfigure[]{\includegraphics[width=0.15\columnwidth]{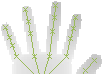}\label{fig:tochange06}}
  \caption[Motivation figure]{(a) A learned joint regressor might fail to 
recover  the pose of a hand due to ambiguities or lack of training data. (b) We 
make use of the inherent uncertainty of a regressor by enforcing it to generate 
multiple proposals. The crosses show the top three proposals for the proximal interphalangeal 
joint of the ring finger for which the corresponding ground truth position is drawn in green. 
The marker size of the proposals corresponds to degree of confidence.
(c) A subsequent model-based optimisation procedure exploits these 
proposals to estimate the true pose. (d) The ground truth for this particular 
example. 
The same colours are used for the corresponding results throughout the work.
}
  \label{fig:motivation}
\end{figure}

The accurate estimation and tracking of hand articulations provides the basis for 
many applications like human computer interaction, activity analysis, and sign 
language recognition.
Hand pose estimation, despite its long history 
\cite{rehg94eccv,Wu01iccv,Rosales06ijcv,Erol07cviu}, 
has attracted more and more interest from the computer vision community in 
recent years \cite{Kyriazis13cvpr,Qian14cvpr,Sharp15chi,Xu15ijcv}. 
This is partly driven by the fact that low-cost sensors can now provide 
reliable depth information.

Most current approaches to tracking of hand articulations can be roughly 
categorized into model-based and data-driven schemes. 
In model-based schemes~\cite{Wu01iccv,Oikonomidis10accv,Gorce11pami,Melax13gi} 
an underlying 3D hand model is used to render pose hypotheses, which are subsequently compared to the observations retrieved from the sensor. 
Since it is infeasible to search the whole range of possible hand poses, these 
methods rely on an initialisation which already needs to be close to the true 
solution. Typically, 
the solution from the previous frame is used for 
initialisation, which leads to problems in the case of very fast hand movements or dropped frames. Hence, subsequent tracking failures are hard to recover from.

On the other hand, data-driven schemes learn the mapping from specific appearances to hand poses from training data~\cite{Tang14cvpr,Keskin12eccv}. 
During testing, they usually infer joint locations independently from each 
other. 
In this way, the complex dependencies do not need to be modelled. However,
the results are not constrained by hand anatomy or physics. Thus, the obtained 
pose estimates might be wrong or even impossible. Another issue for these 
approaches is that employing enough training data to densely cover the whole 
pose space is infeasible, because of the highly articulated nature of the human 
hand and the fact that the space of possible hand poses grows exponentially with 
the number of joints.

In this work we propose a hybrid method with both data-driven and model-based elements that inherits the advantages of both paradigms. 
Very recently, this direction has attracted considerable research (\eg 
\cite{Ballan12eccv,Tang13iccv,Qian14cvpr,Tompson14tog}),
which has been inspired in parts by several relevant and successful approaches on human pose estimation.
In general, there are two types of hybrid approaches. 
Either these approaches try to find a 
solution using the model-based 
approach first, and only consider the data-driven approach for 
failure cases~\cite{Ganapathi10cvpr,Wei12tog}, 
or they obtain (an) initial pose(s) based on the data-driven approach, 
and then validate and/or 
locally optimise the pose(s) using a model-based 
approach~\cite{Baak11iccv,Ye11iccv,Taylor12cvpr}.
While the chosen optimisation scheme, failure detection and runtime 
performance are critical issues for the former approaches, the latter 
ones naturally avoid the computationally difficult problem of global 
optimisation and do not get stuck to local minima.
However, several of the latter approaches either only validate the 
hypotheses generated by the data-driven approach using the model-based method~\cite{Rosales06ijcv,Xu13iccv}, 
or do not consider anatomic constraints. Thus, the resulting poses might be 
invalid~\cite{Rosales06ijcv,Baak11iccv,Ye11iccv}.
Moreover, for the task of 3D hand pose estimation, inherent difficulties like the substantial similarities between individual fingers and the very fast movements or 
complex finger interactions cause ambiguities and uncertainties which are often disregarded by previous works.

\subsection{Exploiting Uncertainties}

For deriving more informed decisions under uncertainty, 
Graphical Models have been proven very effective 
\cite{Fischler73pictstruct,KollerFriedman09PGM,Felzenszwalb05ijcv}.
Hence, they have been extensively used in computer vision literature 
\cite{Geman84pami,He04cvpr,Felzenszwalb10pami}.
Besides being applied in a dense manner 
\cite{He04cvpr,Fulkerson09iccv,Kraehenbuehl11nips}, 
sparse graphical models have become increasingly popular for tasks like object detection or pose estimation~\cite{Wu01iccv,Hamer09iccv,Felzenszwalb10pami,Dantone14pami}.
Despite their effectiveness, a naive implementation would lack efficiency due to the complex interactions 
which need to be modelled. To this end, approximations of the underlying distributions have enabled efficient inference~\cite{Felzenszwalb10pami,Kraehenbuehl11nips,Kraehenbuehl13icml}.

To exploit the inherent uncertainties in the task of 3D hand pose estimation, we first need to capture them.
For this, we adopt successful work on body pose 
estimation~\cite{Girshick11iccv,Shotton13pami} to train a regressor that is able to generate a distribution of location proposals for each joint. We input this distribution to a subsequent optimisation procedure.

Within the optimisation procedure we can then exploit the uncertainties, which are implicitly captured by the proposal distribution.
To do this efficiently we employ an approximation of the full distribution upon which a graphical model operates.
The optimisation procedure considers multiple entirely different solutions for 
the global pose configuration, capturing the 
uncertainty of preceding processing steps.
In this way, the regressor does not need to be perfectly 
accurate on its own but should only deliver a set of likely joint positions, which are subsequently refined.
This also attenuates the need for a complete training database densely covering the whole pose space. 
Additionally, the whole process operates in 3D and is thus able to infer correct joint locations even in the case of occlusions and missing depth information.
Moreover, optimisation not only exploits the uncertainties, but also finds an anatomically valid hand pose, similar to inverse kinematics (see 
Fig.~\ref{fig:motivation}).

\subsection{Related Work}
In previous attempts to apply hybrid approaches specifically to hand pose estimation, the joint proposals provided by the data-driven approach are often refined by 
adding penalties to anatomically implausible joint locations~\cite{Tang13iccv,Poudel13isvc}. 
However, the obtained hand poses can still be invalid since the refinement 
performs only a selection of the most plausible joint proposals and/or refines 
only some of the provided proposals. Hence, the approaches fail 
if all proposals for a single joint are inaccurate (\eg, due to occlusions), or if the proposals are uncertain for many of the joints.

In contrast, our method introduces new joint positions, 
which respect anatomic constraints and, simultaneously, best fit the joint positions proposed by the data-driven approach in a global manner. Moreover, the proposed method does not rely on finding proposals of high confidence, but incorporates the approximated proposal distributions for each joint to find the anatomically valid hand pose which explains them best.

Another strand of research combines salient point detection 
with model-based optimisation~\cite{Qian14cvpr,Ballan12eccv,Tzionas14gcpr}.
However, these approaches rely on detection of specific landmarks and will fail in situations
where the landmarks (usually the finger tips or nails) are not clearly visible.
In contrast, we do not rely on any landmark to be visible, but take a more holistic view considering the whole hand.
Thus, our approach is robust to occlusion of specific landmarks.

Also noteworthy is an approach to 2D body pose estimation~\cite{Dantone13cvpr,Dantone14pami}.
This DPM based work focuses on improving the unaries provided by a Random Forest.
In contrast to their work, we use hard constraints on the graphical model,
which is enabled by defining it in 3D.
However, their ideas for improving the unaries could potentially be applied to our task too.

Probably most closely related to our method is the approach from Tompson~\etal~\cite{Tompson14tog}. 
This approach uses a deep convolutional network to infer the most likely 
positions of some predefined points on the hand which are then optimised by a 
model similar to ours. Despite employing depth information, regression is 
performed solely in 2D, disregarding occlusions or ``holes'' in the depth map. 
Moreover, in contrast to our approach, they rely on a single best location to 
fit the model, which ignores the uncertainty in the regression.

The basic building blocks of our method are similar to those of other hybrid approaches to hand pose estimation:
a discriminative, data-driven method that generates likely joint positions (as in \cite{Tang13iccv,Poudel13isvc}), 
and a generative, model-based optimisation method that refines the initial solution (as in \cite{Qian14cvpr,Tompson14tog}). However, the way we combine these two components is shown to outperform competing approaches. Another key difference that makes our method distinct from any other method we are aware of is that the optimisation component has access to internal information of the data-driven component. It can thus make more informed decisions under the given uncertainty, which again yields significantly improved results.

\section{Method}
In this section we present the two building blocks which make up the proposed method. We use a discriminative regressor, which generates an approximation of the 
proposal distribution (Sec.~\ref{sec:regression}).
This distribution can be effectively transformed to anatomically valid pose hypotheses using the model-based optimisation procedure 
described in Sec.~\ref{sec:modelbasedoptimisation}. 

\subsection{Joint Regression}
\label{sec:regression}
For the generation of an approximated proposal distribution we build upon the 
prominent approach of Girshick~\etal~\cite{Girshick11iccv,Shotton13pami}. 
This approach has been shown to work well for real world applications on body 
pose estimation~\cite{Shotton13pami}, and has also been previously adapted for 
hand pose estimation~\cite{Tang13iccv}. 
The approach relies on Random Forests~\cite{Amit94tr,Breiman01,Criminisi12ft} 
to infer a 3D distribution of likely hand joint locations\footnote{
By joint locations we refer to any annotated positions defining the pose of a hand 
(most often these are joints)}. 
We briefly describe the training and testing procedures as applied in this work since its internal 
information is later exploited during optimisation 
(Sec.~\ref{sec:modelbasedoptimisation}). 
For more details the interested reader is referred to the related 
work.

\noindent{\bf Training:}
For our task we follow a part based approach to learn a mapping $\mathcal{F} : \mathcal{X} 
\rightarrow \mathcal{Y}$.
An input sample $\mathbf{x} \in \mathcal{X} = \mathbb{R}^{D}$ 
represents the local appearance of an image patch around a 
foreground pixel from which we want to infer the 3D locations of $J$ joints, 
\ie, $\mathcal{Y} = \mathbb{R}^{3 \times J}$. 
A training sample is formed by associating the image patch with a part label 
and corresponding offset vectors from the patch centre location to each of the 
joint positions.
Hence, the training set 
$\mathcal{L} = \left\{ (\mathbf{x}_i,c_i,\mathbf{o}_i) \right\} 
\subseteq \mathcal{X} \times \mathcal{C} \times \mathcal{O}$, where $c_i \in 
\mathcal{C} = \left\{1,\ldots,J 
\right\}$
denotes the class label of the joint which is closest to the location of 
$\mathbf{x}_i$, and 
$\mathbf{o}_i \in \mathcal{O} = \mathbb{R}^{3 \times J}$ 
denotes the set of 3D offset vectors. 
The training data is recursively split by each tree individually, 
until the maximum depth of a tree (23 in this work) is reached
or less than a minimum number of 
samples (40) arrives at a node. 
For the experiments we fixed the number of trees to three.

Inspired by~\cite{Schulter11bmvc}, we sub-sample the data arrived 
at a node for split selection. 
This not only speeds up the training process and enforces de-correlation between 
the trees, but also implicitly accounts for the different number of samples 
which are extracted per class by drawing a balanced sub-sample.
The learned split functions are based upon the same simple depth features  
used in \cite{Shotton13pami}.
Finally, to generate the prediction models at the leaves, mean-shift 
\cite{Comaniciu02pami} is applied to the collected offset distributions for each 
joint~\cite{Shotton13pami}.

\noindent{\bf Evaluation:}
\label{sec:regressionusingmeanshift}
During test time we start with an empty set of proposals 
$\mathcal{P}_j = \emptyset$ for each joint $j$. 
Image patches are sampled densely from the foreground region of the depth image 
and are passed down through each tree of the forest. 
Arriving at leaf $l$, the 3D centre position of a test sample 
$\mathbf{x}$ is offset by the stored offset vectors 
$\mathcal{O}^{(l(\mathbf{x}))}_{j}$ for each joint $j$ to form sets of proposals 
$\{ \mathcal{P}^{(l(\mathbf{x}))}_{j} \}$ which are added to the current sets:
$\mathcal{P}_j \leftarrow \mathcal{P}_j \cup \mathcal{P}^{(l(\mathbf{x}))}_{j}$. 
Following~\cite{Shotton13pami} we keep only a reduced set of top confident proposals for each 
joint $\tilde{\mathcal{P}}_j \subseteq \mathcal{P}_j$
and, subsequently, perform mean-shift on those to extract the $k$ top modes 
$\mathcal{P}^{(m)}_{j}$. 
We thus end up with at most $k \times J$ final proposals, each associated with a confidence score.
We base the score on the number of initial proposals supporting 
the mode~\cite{Shotton13pami}. 
Importantly, this set of proposals and confidences approximates the proposal distribution for each joint.

\subsection{Model-based Optimisation}
\label{sec:modelbasedoptimisation}
Using the discriminative Random Forest based method described above, inference of the individual joint proposals is completely independent from the other joints. While, in this way, the complex dependencies do not need to be modelled, the resulting proposals are not necessarily compatible with anatomical constraints.

In order to obtain a valid pose we employ a predefined 3D model of a hand. We use a model with 26 degrees of freedom (DoFs). In this model the global pose of the hand, \ie, 
position and orientation, has six DoFs, and each of the five 
fingers is specified by four more. 
These four parameters per finger encode angles where the base joint of each 
finger is assigned two DoFs and the two remaining hinge joints are each assigned one DoF.
During optimisation these DoFs are constrained 
based on anatomical studies~\cite{Lin00handconstraints,Albrecht03handmodels} 
which avoids impossible configurations.
Since a quaternion representation is used for the global orientation, 
the 26 DoFs are modelled by 27 parameters. 

While the used hand model is, in principle, similar to what is used in related work on hand pose estimation and tracking (\eg, \cite{Oikonomidis11bmvc,Qian14cvpr}), our model only specifies the joint positions instead of specifically designed geometric primitives~\cite{Oikonomidis11bmvc,Qian14cvpr}, or even a 
complete mesh~\cite{Taylor14cvpr,Sharp15chi,Khamis15cvpr}.
As pointed out later in this section, this has very important implications on the 
computational performance of the optimisation process.

Having defined a hand model, the goal is to find the 27 model parameters which best 
describe the modes $\mathcal{P}^{(m)}$ of the joint proposals, obtained from 
the regression forest. To this end, the objective function 
$E ( \mathcal{P}^{(m)},\mathbf{h} )$ judges the quality of any hypothesized 
parameter set $\mathbf{h} \in \mathbb{R}^{27}$. 
More specifically, given a 
function $\delta_j(\mathbf{h})$ which extracts the position of joint $j$ from 
hypothesis $\mathbf{h}$, the objective is formulated as:
\begin{equation}
	E \left( \mathcal{P},\mathbf{h} \right) = \sum_{j=1}^{J}{\max_r \left( w_{j_r} \left( 1 - d_{j_r}^{2} \right) \right)},
	\label{eq:modeloptimisationobjective}
\end{equation} 
where
\begin{equation}
	d_{j_r} = \min \left( 1, \frac{\left\| \mathbf{p}_{j_r} - \delta_j(\mathbf{h}) \right\|_{2}}{d_{max}} \right).
\end{equation} 
Here $\left\|.\right\|_2$ is the $L^2$ norm of the argument, $d_{max}$ is the 
clamping distance, $\mathbf{p}_{j_r} \in \mathcal{P}$ denotes the $r$-th 
proposal for joint $j$, and $w_{j_r}$ is the normalized confidence so that it 
exhibits the properties of a probability. 
Intuitively, the objective enforces those proposals to be selected  
which -- together -- best form an anatomically valid pose.
This selection is guided by the confidence of each proposal.
Moreover, by considering all the top modes of the proposal distribution for a joint, 
we overcome the problem of outlier modes (\eg, proposals for the wrong finger). 
The model will simply converge to joint positions close to those modes 
which best fit into the overall model. This is achieved by optimising the 
objective for the best parameter set $\mathbf{h}^{\star}$: 
\begin{equation}
	\mathbf{h}^{\star} \triangleq \argmax_{\mathbf{h}} \, E \left( \mathcal{P}^{(m)},\mathbf{h} \right).
\end{equation}

It is important to note that the definition of the objective function is based on a small number of 3D distances. As a result, no 3D hand model rendering is required to evaluate the objective function as, \eg in~\cite{Oikonomidis11bmvc,Sharp15chi}. On the contrary, it can be computed very efficiently on conventional (\ie, CPU) processors.

The objective function is optimised by Particle Swarm Optimisation (PSO)~\cite{kennedypso} as employed in many other works on 3D hand pose estimation~\cite{Oikonomidis11bmvc,Kyriazis13cvpr,Qian14cvpr,Tompson14tog,Sharp15chi}. 
PSO performs optimisation by 
evolving a number of particles (solutions) that evolve in parallel over a number 
of generations (iterations). 
If not stated otherwise, an overall number of 50 generations turned out to be sufficient for our experiments.
The incorporated randomness, which is introduced during the 
initialisation of the 
particles as well as during their evolution, makes the method well suited for 
the employed non-convex, non-smooth objective 
function.

\noindent{\bf Stepwise optimisation:}
Any search space grows exponentially with the number of its dimensions. Hence, decomposing the space into non-overlapping sub-spaces offers the possibility of a very significant speed-up. We exploit this for hand pose estimation based on the observation that, given the global orientation of the hand, the fingers can move almost independently of each other. This independence together with the performed mapping of the proposals to specific joints allows us to split the 
optimisation problem into sub-problems of lower dimensionality. In line with this, we treat the problem of finding the best 27 parameters as six sub-problems, where we first optimise for the 7 parameters specifying the global pose of the palm, and subsequently individually optimise for the 4 parameters of each finger.

\section{Experiments}
We prove the applicability of the proposed method by means of several experiments on synthetic as well as real world data. 
A crucial issue for benchmark datasets is the availability of ground truth.
However, accurate 3D annotations for real data are not easily obtained, especially for articulated self occluding objects like the human hand. To overcome this issue, we employ synthetic data in addition to real data.
To generate a synthetic sequence which resembles natural movements, a 
hand is tracked with the method described in~\cite{Oikonomidis11bmvc} 
using the publicly available implementation\footnote{publicly available at 
\url{http://cvrlcode.ics.forth.gr/handtracking/}} and with a 
very high computational budget. 
We then render depth maps from the resulting poses and utilize the renderings as the test sequence.
In this synthetic sequence the hand performs various finger articulations and typical motions like counting, pinching
and grasping.
While the performed movements are slow to ensure 
that the tracker does not get lost, we afterwards 
sampled every 5th frame from the sequence to simulate a more natural speed of movements. 
This sequence is referred to as \emph{TrackSeq}.

For producing training data, we first defined 4 different articulations per finger. All 1024 combinations of these articulations were used as an initial set of poses. These poses were then rendered under 7 different viewpoints to create the full train set\footnote{Note, that rendering from a different viewpoint is equivalent to changing the orientation of the whole hand.}
of 7168 poses.

For experiments with real data we employ 
the \emph{ICVL Hand Posture Dataset}\footnote{publicly available at 
\url{http://www.iis.ee.ic.ac.uk/~dtang/hand.html}} and \emph{NYU Hand Pose 
Dataset}\footnote{publicly available at 
\url{http://cims.nyu.edu/~tompson/NYU_Hand_Pose_Dataset.htm}}, 
where we use the available training and test data as is.
The ICVL dataset was acquired using the Intel Creative 
\emph{Time-of-Flight} (ToF) camera and includes a training set 
with roughly 330K images and two test sequences with 702 and 894 images, 
respectively.
The two test sequences show a hand facing towards the camera performing various finger articulations in very fast succession.
The NYU dataset was acquired using the Kinect RGB-D camera 
and includes a training set with roughly 73K images and a test set 
capturing two actors and consisting of 8252 images (2440 and 5812, resp.).
However, neither our method, nor that of Tompson~\etal~\cite{Tompson14tog}, 
who published the dataset, can yield any meaningful result for the second actor.
This is because there is only a single actor included in the training set
and the hand of the second actor in the test set differs significantly.
Thus, we only compare on the sequence of the first actor.

\subsection{Influence of Major Processing Steps}
\label{sec:experimentsParameters}
\begin{figure}
  \centering
  
\subfigure[]{\includegraphics[width=0.40\columnwidth]
{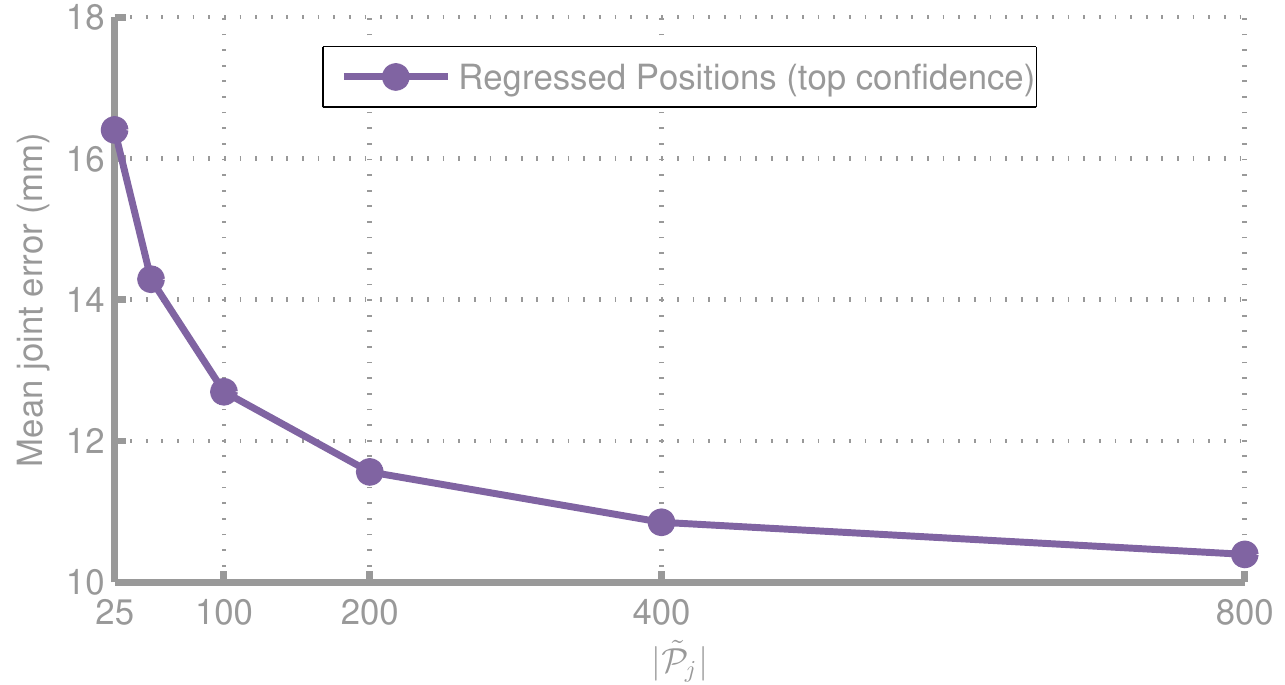}\label{fig:nummeanshiftpoints}}
  \hfil
\subfigure[]{\includegraphics[width=0.40\columnwidth]
{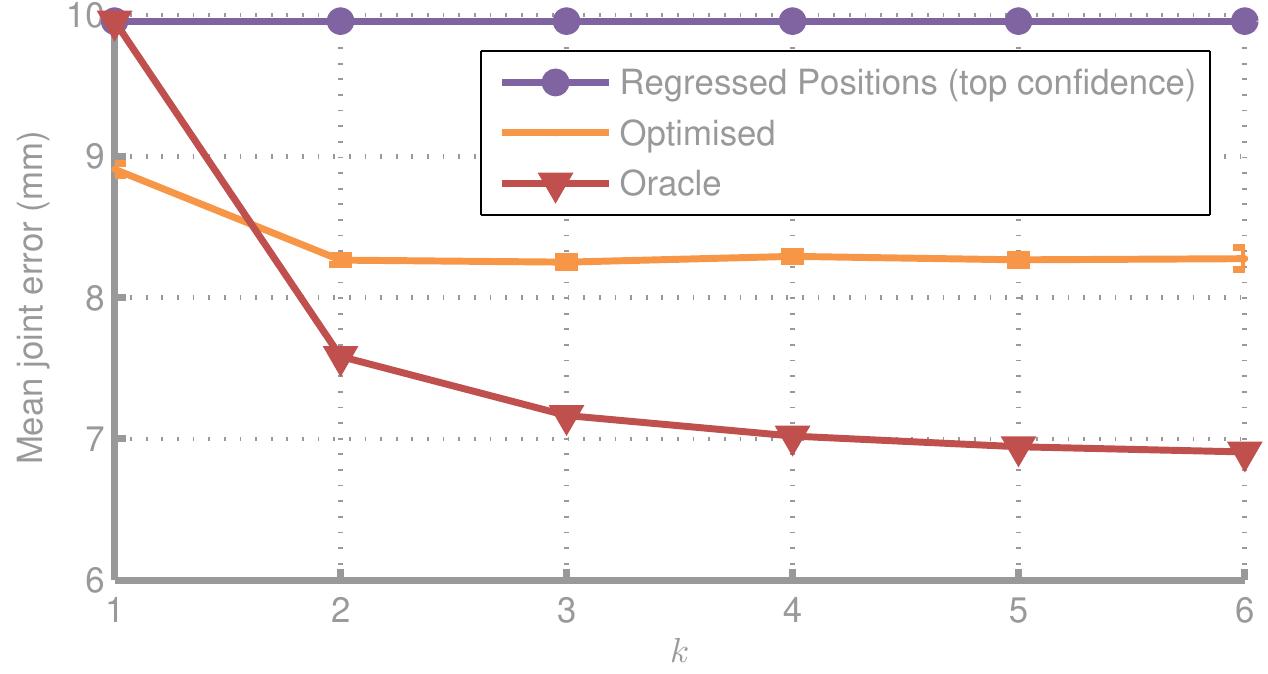}\label{fig:ResultsWrtNumProposals}}
  \caption[Effect of parameter choices]{(a) Mean joint localization error on ICVL dataset as a function of the number of proposals, $\left| \tilde{\mathcal{P}}_{j} \right|$, which are input to mean-shift to extract the final joint proposals at test time. (b) Mean joint localization error on the \emph{TrackSeq} sequence as a function  of the number of top proposals, $k$. 
For \emph{Optimised} the error bars show the standard deviation over multiple runs.
	} 
  \label{fig:forestparameters}
\end{figure}

\noindent {\bf Size of proposal sets $\tilde{\mathcal{P}}_{j}$:}
An important parameter of our method is the number of proposals, 
$\left| \tilde{\mathcal{P}}_{j} \right|$, 
which are input to mean-shift at test time (see 
Sec.~\ref{sec:regressionusingmeanshift}). 
This is especially interesting since 
mean-shift is responsible for summarizing the proposal distribution that we want 
to exploit during optimisation.
As can be seen from Fig.~\ref{fig:nummeanshiftpoints}, the gain in 
accuracy becomes smaller for a higher number of proposals. 
This enables us to find a good trade-off between speed and accuracy. 
In our case, we fix $\left| \tilde{\mathcal{P}}_{j} \right| = 200$ for all other experiments.
Another interesting observation is that accuracy seems to level off much later than reported 
for body pose estimation in \cite{Shotton13pami}.
We hypothesize that this difference is due to the higher variation of hand poses 
compared to body poses within the respective datasets.
In any case, it further advocates the specific consideration of the 
uncertainty as proposed in this work.

\noindent {\bf Number of final proposals per joint:}
The optimiser can efficiently exploit the inherent uncertainty of the regression process due to the approximation of the proposal distribution for each joint. 
Thus, we investigate the effect of the number of generated proposals $k$ on accuracy.
Fig.~\ref{fig:ResultsWrtNumProposals} shows the error with respect to $k$.
The \emph{Oracle} always selects the proposal closest to the respective ground truth joint position. Obviously, the more proposals generated, the closer one  of them will be to the ground truth. The error for the proposed method (\emph{Optimised}) is higher because the solution has to respect the anatomical constraints of the hand. Interestingly,  accuracy levels off for a small number of proposals (2-3) both for \emph{Oracle} and for \emph{Optimised}.
Hence, the results show that utilizing a small number of proposals (together with confidences) 
instead of the full proposal distribution $\mathcal{P}^{(i)}$ 
is already very effective.
For the other experiments we thus set $k = 3$.
In fact this also limits the complexity of the objective function 
(Eq.~\eqref{eq:modeloptimisationobjective}).

Furthermore, Fig.~\ref{fig:ResultsWrtNumProposals} indicates that the results of plain 
regression can already be improved by applying optimisation to the single top proposal for each joint. This improvement can be attributed to the anatomically valid solution induced by model-based optimisation.
As also suggested by the results (for $k=1$ and $k=3$) in Fig.~\ref{fig:resultsForthIcvl},
a significant additional gain is achieved by providing the optimisation procedure with internal information about the uncertainty of the regressor, \ie, multiple proposals per joint.

\noindent{\bf Stepwise optimisation:}
We investigate the effect of the stepwise optimisation procedure described in Sec.~\ref{sec:modelbasedoptimisation}.
For a meaningful comparison we fix the overall number of objective function evaluations 
(\ie, the optimisation budget) for both approaches.
We used 91 particles and generations when optimising all parameters together, 
whereas for the stagewise optimisation
we assigned 64 particles and generations to the optimisation of position and orientation 
and 29 particles and generations to optimisation of each finger.
The results are shown in Fig.~\ref{fig:ratioframeswithinthreshTrackSeq} (orange curves). 
Despite the relatively high optimisation budget, 
the procedure does not converge when optimising all parameters together.
That is, 
the stepwise optimisation procedure proves much more effective.

\noindent {\bf Runtime:}
Our current implementation of the Random Forest based regressor 
takes $\sim$\unit[33]{msec} on an Intel i7-4820K CPU 
to compute the joint proposals which are input to the optimiser.
Optimisation itself obviously depends on the budget.
For the current implementation 1000 objective function evaluations 
take $\sim$\unit[10]{msec}.
If not stated otherwise, we fixed the number of objective evaluations 
to $\sim$3400 for all experiments for a good speed vs. accuracy trade-off.
Note that all timings are given for our current prototype implementation, 
which is non-optimised and \emph{single} threaded.

\subsection{Comparison to the State of the Art}
\label{sec:experimentsSota}
We compare our method to state of the art 
model-based (FORTH, \cite{Oikonomidis11bmvc}), 
data-driven (LRF, \cite{Tang14cvpr}) and 
hybrid (Tompson~\etal, \cite{Tompson14tog}) approaches.
The LRF method has recently been shown to outperform other  
data-driven as well as model-based approaches~\cite{Tang14cvpr}.
Note that, all our results can be found on our website\footnote{Results and other material can be found at 
\url{http://lrs.icg.tugraz.at/research/hybridhape/}}.

\begin{figure}
  \centering
  \subfigure[]{\includegraphics[width=0.28\columnwidth]{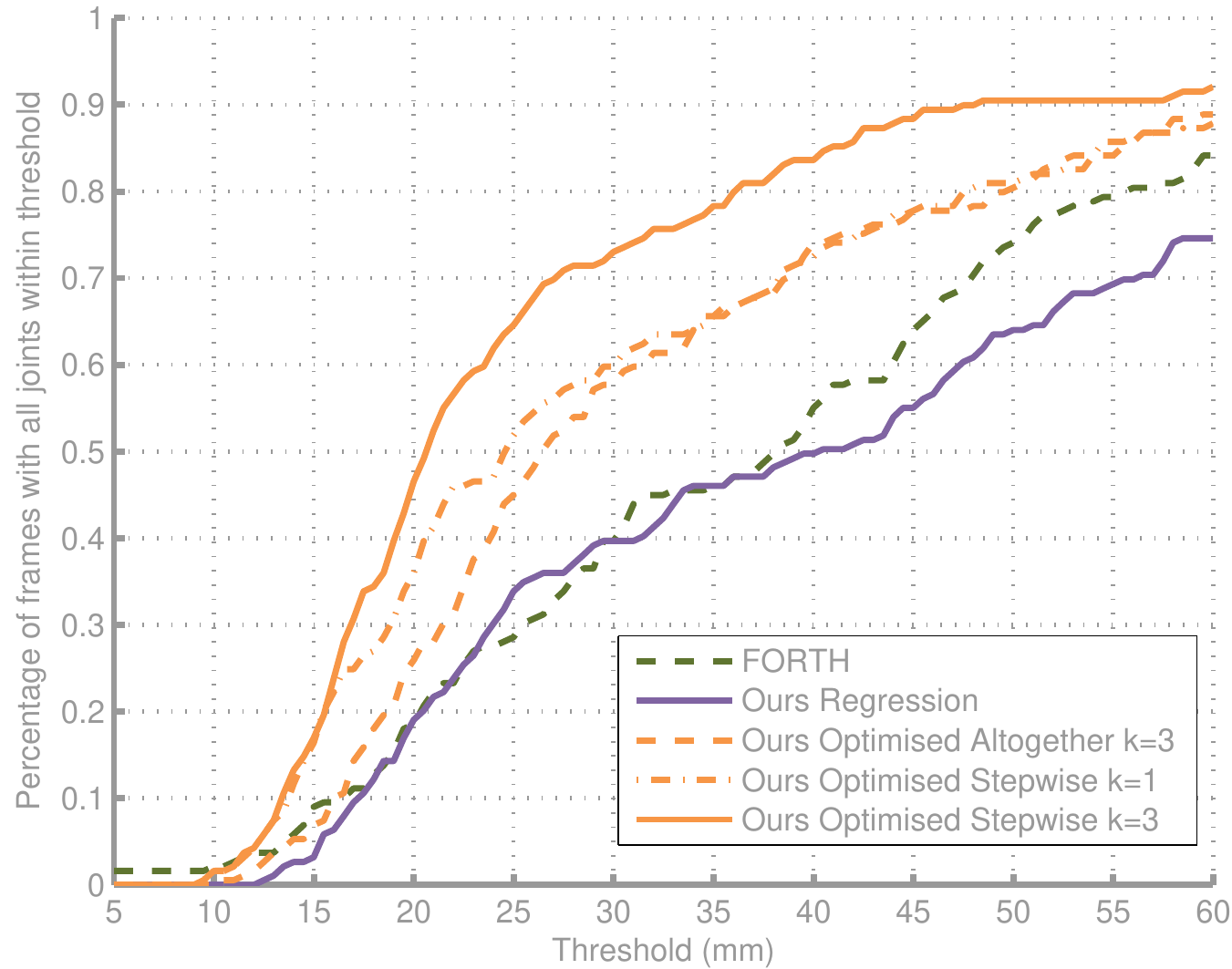}\label{fig:ratioframeswithinthreshTrackSeq}}
  \hfil
  \subfigure[]{\includegraphics[width=0.28\columnwidth]{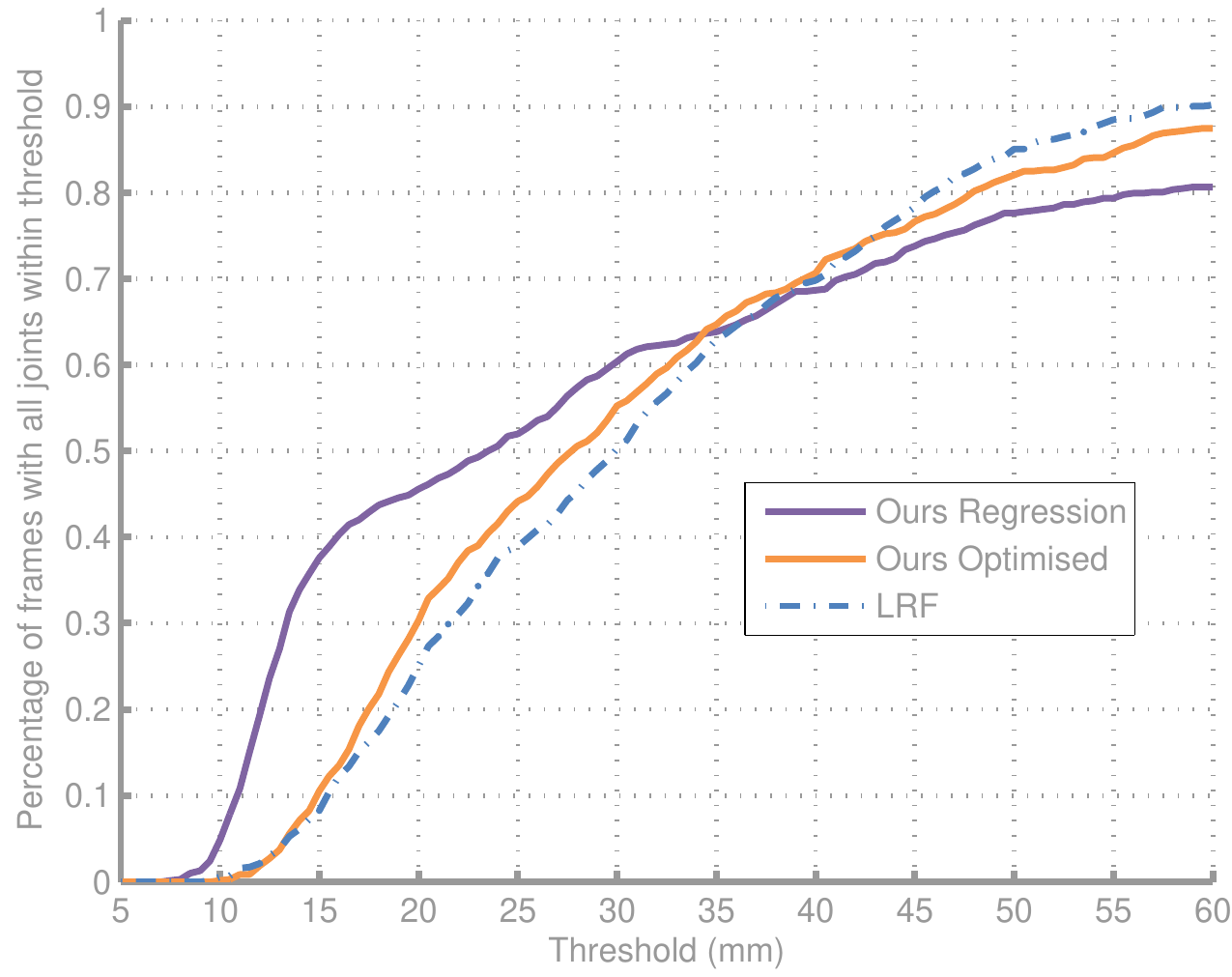}\label{fig:successRateFramesIcvl1}}
  \hfil
  \subfigure[]{\includegraphics[width=0.28\columnwidth]{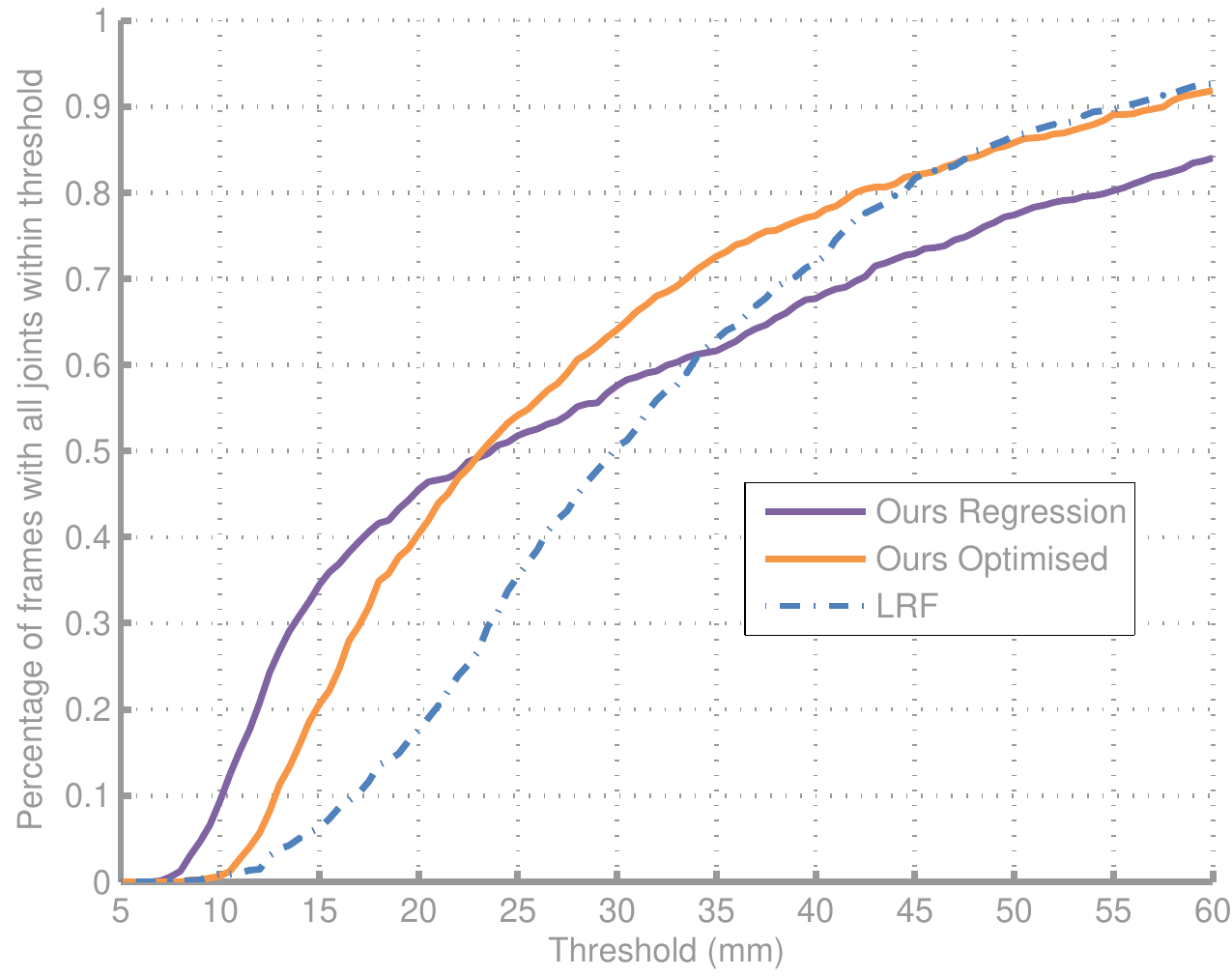}\label{fig:successRateFramesIcvl2}}
  \caption[Ratio frames within threshold on TrackSeq and ICVL test sequences]{
	Success rate as a function of the distance threshold for (a) the 
	\emph{TrackSeq} test sequence, (b) the ICVL dataset 
	\emph{Sequence~1} and (c) the ICVL dataset \emph{Sequence~2}.
	It shows the ratio of frames with \emph{all} joints within a certain distance to the ground truth 
	as a function of this distance.
	Results are compared to those from \emph{FORTH}~\cite{Oikonomidis11bmvc} 
	and \emph{LRF}~\cite{Tang14cvpr}.} 
  \label{fig:resultsForthIcvl}
\end{figure}

\noindent {\bf Proposed vs FORTH~\cite{Oikonomidis11bmvc}:} 
 We evaluate our method on the \emph{TrackSeq} test sequence.
Fig.~\ref{fig:ratioframeswithinthreshTrackSeq} compares 
the frame-based \emph{success rate} over a number of distance thresholds.
The frame-based success rate gives the ratio of frames in which all joints are estimated 
within a certain threshold to ground truth.
The results show that our plain regressor performs similarly to \cite{Oikonomidis11bmvc}
for the most interesting range of thresholds. 
However, after enforcing anatomic constraints by model-based optimisation, the proposed hybrid method is able to improve on these results by a large margin.
This is achieved despite the fact that the sequence exhibits strong finger articulations, whereas the position and orientation do not change.
Thus, the main reason for this gain  
is the improved estimation of finger articulations 
rather than the overall position and orientation estimation.
This is also illustrated when solely considering the errors in finger tip 
localization, where the mean error for~\cite{Oikonomidis11bmvc} is 19.5mm, 
while for our approach it is 11.8mm -- an error reduction of about 40\%.

\noindent {\bf Proposed vs LRF~\cite{Tang14cvpr}:} 
We compare to LRF~\cite{Tang14cvpr} on the ICVL dataset.
To ensure a fair comparison we use their results published online.
As shown in Fig.~\ref{fig:successRateFramesIcvl1} and \ref{fig:successRateFramesIcvl2} 
our regressor outperforms their results over most of the thresholds by a significant margin.

Unfortunately, we cannot fairly evaluate our model-based optimisation using the annotations provided with the ICVL dataset.
This is because in the ground truth annotations, bone lengths\footnote{By bone lengths we refer to the distance between joint annotations connected by bones} 
vary significantly between the frames of a single sequence; therefore they are not compatible with an anatomically valid  3D hand.
However, the 26 DoFs hand model used in this work implicitly applies strict constraints on them.
Fitting such a model will therefore always introduce additional errors when compared 
to the provided ground truth annotations.
Nevertheless, our results appear more accurate or at least as accurate as those from LRF~\cite{Tang14cvpr}.

\begin{figure}
  \centering
  \subfigure[]{\includegraphics[width=0.28\columnwidth]{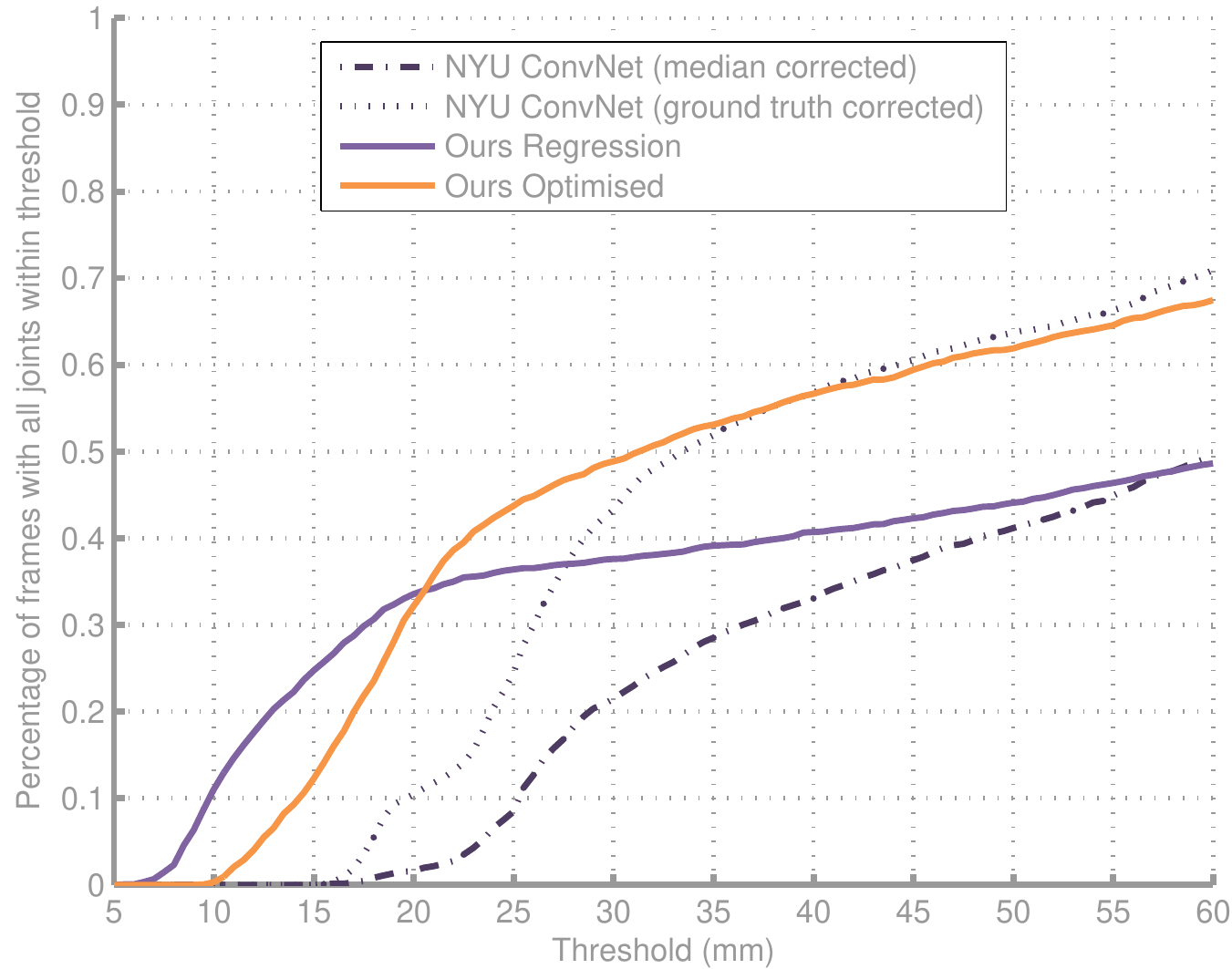}\label{fig:nyuSuccessRateFramesActor1}}
  \hfil
  \subfigure[]{\includegraphics[width=0.59\columnwidth]{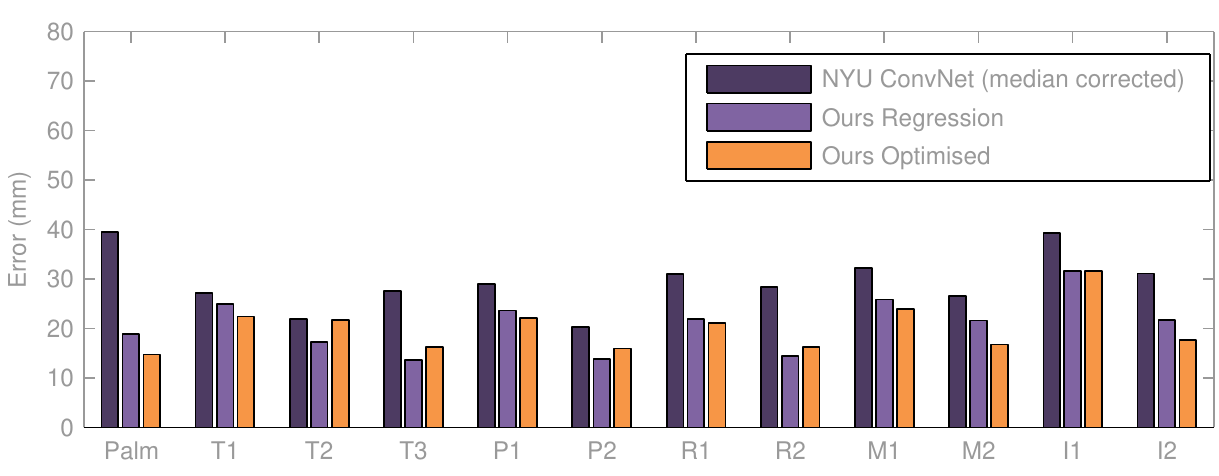}\label{fig:errorPerJointActor1}}
  \caption[Results on NYU dataset (\emph{Actor~1})]{
	Results for the NYU dataset.
	(a) The ratio of frames where \emph{all} inferred positions are within a certain distance from ground truth as a function of this distance. (b) The average error per joint.
	Results are compared to the hybrid approach 
	of~\cite{Tompson14tog}
	(\emph{NYU ConvNet}).
	} 
  \label{fig:resultsNyu}
\end{figure}

\noindent {\bf Proposed vs Tompson~\cite{Tompson14tog}:} 
We compare to the hybrid approach of Tompson~\etal~\cite{Tompson14tog} 
based on the NYU dataset. 
For this dataset the annotated positions do not actually correspond to joints, but to some specific locations on the hand.
We used the suggested positions for evaluation~\cite{Tompson14tog} 
with the minor exception that we had to skip two of the three palm positions 
since the palm is only represented by a single position in our model. 
In addition, \cite{Tompson14tog} provides only 2D locations, 
which are projected to 3D using the input depth. 
However, the input depth might be significantly distorted (\eg, at holes). 
To correct inferred locations, which lie at positions with distorted depth,
we augment them with the median depth of the inferred positions in the same frame.
Fig.~\ref{fig:resultsNyu} shows the results and also compares to a variant 
where we correct the depth using the ground truth depth of the corresponding joints 
to show the theoretical upper bound for their approach. 
For our method the optimisation improves results particularly for larger distance thresholds 
since the optimisation mainly performs an error correction rather than improving inferred locations,
which are already very close to ground truth. 
In addition, the difference between the annotation model and our model induces an error, especially for low thresholds.
In spite of that, we observe that our method outperforms the approach of 
Tompson~\etal~\cite{Tompson14tog} by a large margin -- 
even if we correct their results by augmenting ground truth depth information.

\section{Conclusions}
We proposed a hybrid approach for 3D hand pose estimation based on a single depth frame. A regression forest delivers several proposals for each hand joint position. Then, model-based optimisation is responsible for estimating the best fit of a 3D hand model to the joint proposals obtained through regression. Thus, optimisation  exploits the inherent uncertainty of the data-driven regression.
As a result, the proposed method delivers anatomically valid solutions (which most 
purely data-driven methods fail to provide) without unrecoverable track losses or a need 
for proper initialization (as happens with most purely model-based approaches). 
The proposed method has been shown to achieve state-of-the-art performance. 
This is proven by quantitative experiments on several datasets and in comparison with representative methods from all categories (model-based, data-driven, hybrid).

\begin{sloppypar}
\paragraph{Acknowledgement}
We thank Axel Pinz for his efforts in reviewing the paper prior to submission, 
and gratefully acknowledge the support and inspiring discussions with 
Paschalis Panteleris, Nikolaos Kyriazis and Iason Oikonomidis 
from the CVRL lab at FORTH and Ren\'{e} Ranftl from ICG.
This work has been supported by the European Union under the Seventh 
Framework Programme, projects 3D-PITOTI (ICT-2011-600545), 
FP7-IP-288533 Robohow and FP7-ICT-2011-9 WEARHAP, 
and by the Austrian Research Promotion Agency (FFG) under the FIT-IT Bridge programme, project TOFUSION (838513).
\end{sloppypar}

\bibliography{_abbrv_,references}

\end{document}